\documentclass[letterpaper]{article} % DO NOT CHANGE THIS
\usepackage{aaai2027}  % DO NOT CHANGE THIS
\usepackage[hyphens]{url}  % DO NOT CHANGE THIS
\usepackage{graphicx} % DO NOT CHANGE THIS
\usepackage{natbib}  % DO NOT CHANGE THIS AND DO NOT ADD ANY OPTIONS TO IT
\usepackage{caption} % DO NOT CHANGE THIS AND DO NOT ADD ANY OPTIONS TO IT
\usepackage{amsmath}
\usepackage{amssymb}
\usepackage{amsthm}
\usepackage{booktabs}

\usepackage{algorithm}   % re-load to restore the algorithm float environment
\usepackage{algpseudocode}
\usepackage[table]{xcolor}
\usepackage{siunitx}
\usepackage{arydshln}
\usepackage{microtype}
\usepackage{nicefrac}
\usepackage{soul}

\nocopyright

\newcommand{\red}[1]{{\color{red} #1}}
\newcommand{\cyan}[1]{{\color{cyan} #1}}

\newtheorem{example}{Example}

\newcommand{\bx}{{\bf x}}

\newcommand{\bz}{{\bf z}}

\newcommand{\ta}{\theta}

\newcommand{\cN}{{\mathcal N}}

\newcommand{\cE}{{\mathcal E}}

\newcommand{\cL}{{\mathcal L}}

\newcommand{\benum}{\begin{enumerate}}
\newcommand{\eenum}{\end{enumerate}}

\newcommand{\bitem}{\begin{itemize}}
\newcommand{\eitem}{\end{itemize}}

\newcommand{\bcode}{\begin{lstlisting}}
\newcommand{\ecode}{\end{lstlisting}}

\newcommand{\bgal}{\begin{align}}
\newcommand{\edal}{\end{align}}

\newcommand{\be}{\begin{equation}}
\newcommand{\ee}{\end{equation}}
\newcommand{\ba}{\left[ \begin{array}{cc}}
\newcommand{\ea}{\end{array} \right]}

\newcommand{\bea}{\begin{eqnarray}}
\newcommand{\eea}{\end{eqnarray}}

\newcommand{\bdf}{\begin{definition}}
\newcommand{\edf}{\end{definition}}
\newcommand{\bex}{\begin{example}}
\newcommand{\eex}{\end{example}}

\title{Follow the Herd: A Data-Driven Approach for Actionable Recourse}
\author{
    Nishtha Madaan\thanks{Nishtha Madaan is a researcher at IBM Research and this work was done as a part of PhD research at IIT Delhi.},
    Srikanta Bedathur\\
}
\affiliations{
    Indian Institute of Technology Delhi, India\\
    nishtha.madaan@iitd.ac.in, srikanta@cse.iitd.ac.in
}

\begin{document}

\maketitle

\begin{abstract}
As machine learning systems make increasingly greater decisions about people's lives, it is critical to provide users with actionable recourse about how they can achieve their desired outcomes. This problem is currently tackled by relying on expensive human supervision which involves hand-designing a cost function before searching for low-cost actions that achieve the desired outcome. In this paper, we propose \textit{``Data-Driven Action Recourse"} (DDAR), the first approach for providing action recourse without requiring costly supervision from humans about action costs. We build upon the notion that highly likely trajectories in the historical user data should be associated with lower costs. With this, DDAR first builds a dynamics model of historical user trajectories. Under this model, it then searches for trajectories that achieve the desired outcome. In experiments, we find that this simple approach recommends plausible action recourse without requiring human supervision, achieving lower action costs than the existing approaches.
\end{abstract}

\section{Introduction}
% High-Level Problem Description

Machine Learning systems are widely adopted in decision-making applications such as criminal justice, loan approval, and healthcare. It is often required to provide the users of these models with a recourse in the case of an unfavorable outcome. For instance, if a user seeking a loan gets her application rejected, she would need a plausible recourse that helps her get the loan application accepted. Therefore, providing plausible and actionable recourse is critical in such scenarios.

% What is the promising field? What is the general framework of action recourse.
The closest line of work pursuing this is that of action recourse \citep{wachter2017counterfactual, mothilal2020explaining, ustun2019actionable}. Broadly, in this line of work, we are given a decision model and a cost function that evaluates how costly an action is. Then, the action recourse problem is formulated as a constrained optimization problem where, given a user state, the aim is to perturb the input state in order to cross the decision boundary of the given decision model to achieve the desired outcome while also incurring the lowest cost to the user during perturbation.

% What are the existing works and how they fail to tackle this problem
The existing works that focus on generating actionable recourse \citep{ustun2019actionable, Rawal2020BeyondIR} rely on human supervision. For example, specifying which attributes are mutable (e.g., income) and which attributes are not (e.g., gender or race) is done by a human---either explicitly as a constraint or implicitly via a cost function. Furthermore, there may be an available recourse which, although actionable, might be less typical or commonly associated with high cost. This kind of common-sense knowledge needs significant human effort to incorporate into the cost functions of existing action recourse approaches. On the other hand, many domain-specific constraints and common-sense knowledge can be automatically derived from the historical data. To the best of our knowledge, no prior work has tackled the question of how to acquire the cost associated with actions in an unsupervised manner from data.

\begin{figure*}
    \centering
    \includegraphics[width=0.98\linewidth]{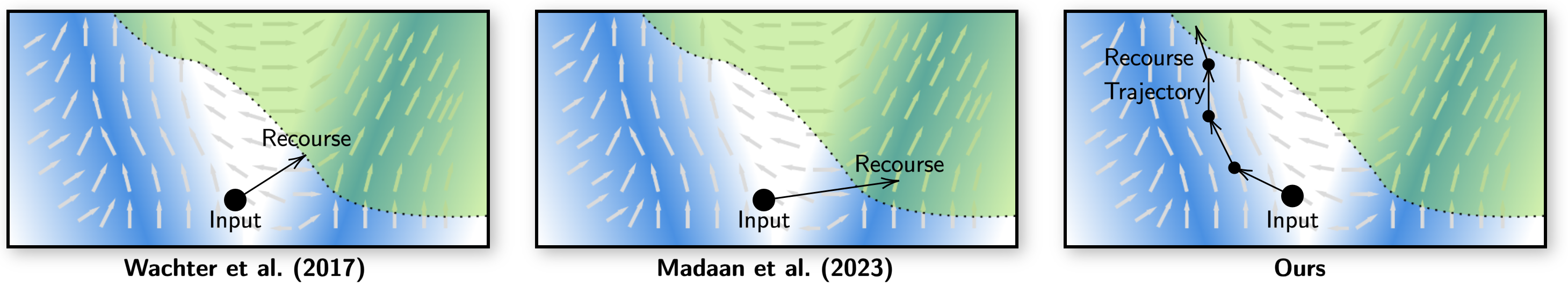}
    \caption{\textbf{Overview.} In this figure, we illustrate the space of user states. The intensity of blue illustrates how likely the user state is in the population. The grey arrows denote the most likely course or direction that users pursue at the given user state. We equate the most likely direction with the direction of a low-cost recourse. The shaded green area denotes the region of state space that achieves the desired outcome from the decision-making system. \textit{Left:} We illustrate that the original formulation of Wachter \textit{et al.} \citep{wachter2017counterfactual} aims only for the nearest recourse, disregarding the actionability and cost of the recourse. \textit{Center:} We illustrate that although the formulation of Madaan \textit{et al.} \citep{madaan2024navigating} emphasizes plausibility when recommending recourse, however, it ignores the cost to the user performing the recourse. \textit{Right:} In contrast to both these previous paradigms, our framework takes into account the recourse cost conditioned on the current state of the individual when recommending a recourse. Furthermore, the notion of the recourse cost is acquired in a data-driven manner without relying on costly human supervision.}
    \label{fig:overview}
\end{figure*}

% How is our work capable to handle this problem?
In this work, we propose a framework called \textit{``Data-Driven Action Recourse"} (DDAR), the first action recourse framework that learns a notion of plausiblity in a data-driven manner. This framework does not require any human supervision. We implement this using a diffusion model trained on historical logs of user trajectories to predict future user states given the current state. Given the trained diffusion model, we generate possible recourse trajectories that are actionable and have low cost for the user. We sample the next possible recourse so that the recommended user state achieves the desired outcome with respect to the decision-making system, but crucially, this process is guided by a dynamics model of user trajectories.

The contributions of this paper are as follows: 1) We introduce DDAR, the first data-driven action recourse framework that learns to generate recourse by acquiring the underlying data distribution and its characteristics, 2) In experiments, we demonstrate that DDAR outperforms the baselines significantly in various metrics. That is, we show that our generated recourse are low-cost and exhibit high recourse rate as compared to baselines. 3) We further extend this framework to generate a $K$-step recourse.

\section{Data-Driven Action Recourse (DDAR) Framework}
In this section, we introduce our proposed framework \textit{Data-Driven Action Recourse} (DDAR).

\textbf{Action Recourse Problem.} In action recourse problem, the aim is to take the previous user state $x_\text{prev}$ and recommend a next user state $x$ which minimizes the cost that the user incurs while also crossing the decision boundary. Formally, the action recourse problem may be described as:
\begin{align*}
    \arg \min_x \text{cost}(x|x_\text{prev}) && \text{s.t. } f(x) = y^*
\end{align*}
where $f$ is a decision model and $y^*$ is the desired label that the user seeks from the decision model $f$.

\textbf{A Data-Driven Cost Function.} It is a challenge to know what the cost function is. Currently, it requires an expensive manual process to define it which is not ideal. Therefore, it is desirable to acquire the cost function in a data-driven manner. We look at the aggregate behavior of users in a similar situation $x_\text{prev}$ and build upon the notion that the most likely next states would be those that correspond to the low cost actions \citep{von2007theory}. This allows us to approximate the cost function as follows:
\begin{align*}
    \text{cost}(x|x_\text{prev}) \propto -\log p_\text{user}(x|x_\text{prev})
\end{align*}
where, $p_\text{user}(x|x_\text{prev})$ is the probability of the next user state $x$ given the current user state with respect to the observational user trajectory data.

\textbf{DDAR Objective.} With this, we obtain our novel and proposed formulation of the data-driven action recourse problem:
\begin{align*}
    \arg \max_x \log p_\text{user}(x|x_\text{prev}) && \text{s.t. } f(x) = y^*
\end{align*}

\textbf{User Model.} We shall refer to the model $p_\text{user}(x|x_\text{prev})$ as the \textit{user model}. We assume that this user model is capable of generating samples and evaluating the density depending on the specifics of the optimization approach used to solve the DDAR objective.

\textbf{Achieving Desired Outcome.} The constraint in the DDAR objective is about whether $x$ has achieved the desired label $y^*$ or not. This constraint can be equivalently expressed as a score or a reward function i.e., a higher score or reward points to the higher likelihood of crossing the decision boundary. 

\textbf{Optimization Approach.} In our general framework, one is free to employ any optimization approach \citep{garey1979computers, metropolis1953equation, geman1984stochastic, browne2012survey, kocsis2006bandit, rubinstein1999cross} given the user model, decision model, and a score function. A naive optimization approach to solve the action recourse problem can consist of the following steps: \textit{1)} sample several possible future trajectories from the user model given the current user state, \textit{2)} score each trajectory and \textit{3)} rank the trajectories and select those that achieve the desired label, and output them as a viable recourse. 

\begin{figure*}
  % optional: fine‐tune vertical placement
  \centering
  \includegraphics[width=0.85\textwidth]{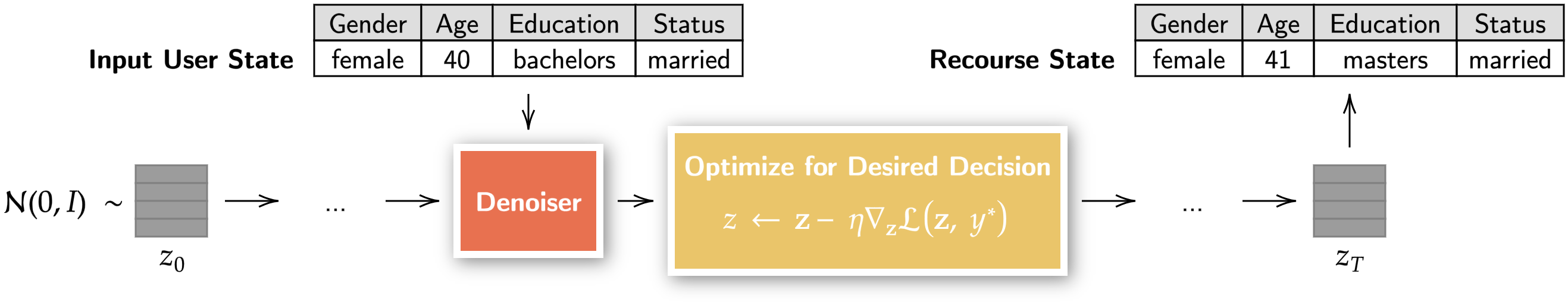}
  \caption{\textbf{Architecture.} In DDAR, the input user state is first encoded and then passed to a denoising diffusion model to produce a recourse embedding that is decoded to produce the final recourse. Crucially, the denoising iterations are intercepted by a gradient‐update step with respect to the decision model, to guide the denoising process towards achieving the desired label.}
  \label{fig:DDAR}
\end{figure*}

\section{Conditional Guided Diffusion for DDAR}

In this section, we describe our proposed approach to implement the DDAR framework. Our implementation rests on two key design choices: (1) We implement our user model via a diffusion model. Diffusion modeling via DDPM \citep{ho2020denoising} has been shown to be an effective yet simple approach for modeling complex distributions. (2) We optimize for the score function by adopting a common approach called \textit{guided-diffusion} given the trained diffusion model. We now describe our approach in detail.

\subsection{Encoder}
\label{sec:encoder}
The encoder $\text{Encoder}_\ta(\cdot)$ maps the raw user state to a latent representation. We assume that the user state at time $l$ is provided as a $K$-sized tuple of integers i.e., $\bx_l \in \mathbb{I}^K$. For instance, if the dataset is tabular and the user state at a certain year $l$ is given as a row in the table with $K$ columns, then, we can quantize each cell to an integer. Then for each $k$, we maintain a learned embedding dictionary $\mathcal{E}_k:\mathbb{I}\rightarrow \mathbb{R}^D$ that maps the integer value to an $D$-sized embedding. By applying embedding dictionaries to the $K$ items in the tuple, we get an encoding of the user state as follows:
\begin{align*}
    \bz_{l} = \text{Encoder}_\ta(\bx_l), && \text{where }\bz_t \in \mathbb{R}^{K\times D}.
\end{align*}

\subsection{Conditional Diffusion as the User Model}
Via diffusion modeling, we seek to learn a distribution $p_\ta(\bz_{l+1} |\bz_{l})$ that takes a current user state $\bz_l$ and predicts the next user state $\bz_{l+1}$. With diffusion modeling, the distribution $p_\ta(\bz_{l+1} |\bz_{l})$ shall consists of $T$ denoising steps, with each denoising step $p_\ta(\bz_{l+1, 0}|\bz_l)$ formulated as:
\begin{align*}
    \int p(\bz_{l+1,T}) \prod_{t=T}^1 p_\ta(\bz_{l+1, t-1}|\bz_{l+1,t}, \bz_l, t) d \bz_{l+1, 1:T}.
\end{align*}
Here, $p(\bz_{l+1, T})$ is a standard Gaussian, the sequence $\bz_{l+1, T}, \ldots, \bz_{l+1,1}$ consists of iteratively cleaner samples, which finally produce the desired sample $\bz_{l+1,0}$. We will use $\bz_{l+1,0}$ and $\bz_{l+1}$ interchangeably as they refer to the same variable. The $p_\ta(\bz_{l+1, t-1}|\bz_{l+1,t}, \bz_l, t)$ is a one-step denoising distribution parameterized as $\cN\left( \gamma_{1,t}\hat{\bz}_{l+1,0} + \gamma_{2, t}\bz_{l+1,t}, \beta_t\mathbf{I}  \right)$, where $\hat{\bz}_{l+1,0} = g_\ta(\bz_{l+1,t}, \bz_{l}, t)$, and the coefficients $\gamma_{1,t}={\beta_t \sqrt{\bar{\alpha}_{t-1}}}/{1 - \bar{\alpha}_t}$ and $\gamma_{2,t} = {(1 - \bar{\alpha}_{t-1})\sqrt{\alpha_t}}/{1 - \bar{\alpha}_t}$.
Employing standard notations, we utilize a variance schedule $\beta_1, \ldots, \beta_T$, where $\alpha_t = 1 - \beta_t$, and $\bar{\alpha}_t = \prod_{i=1}^t (1 - \beta_i)$. We use a cosine schedule in our implementation.

\textbf{Implementing the Denoiser.} We implement the denoising predictive mean $g_\ta(\bz_{l+1,t}, \bz_{l}, t)$ as a transformer. Specifically, we first concatenate $\bz_{l+1,t}$ and $\bz_{l}$ followed by adding learned positional embeddings and step-embeddings representing $t$. The result is provided as input to the transformer followed by an 2-layer MLP to obtain the denoised prediction. We use a 12-layer transformer with hidden size 768 and 12 heads.

\textbf{Training Objective.} Following DDPM, the training procedure involves first introducing noise to the input $\bz_{l+1,0}$, creating its noisy version $\bz_{l+1,t}$ i.e., $\bz_{l+1,t} = \sqrt{\bar{\alpha}_t} \bz_{l+1,0} + \sqrt{1 - \bar{\alpha}_t} \boldsymbol{\epsilon}_t$, where $\boldsymbol{\epsilon}_t \sim \mathcal{N}(\mathbf{0}, \mathbf{I})$.
Subsequently, the denoiser is trained that takes $\bz_{l+1,t}$ as input and aims to predict the original input $\bz_{l+1,0}$ by generating a prediction $\hat{\bz}_{l+1,0} = g_\ta(\bz_{l+1,t}, t)$.
The learning objective is $\cL_\text{diffusion}(\ta) = \cE(\hat{\bz}_{l+1,0}, \bz_{l+1,0})$ where $\cE$ is an error function. As our error function, we use a cross-entropy loss with respect to clean categorical labels corresponding to $\bz_{l+1,0}$. We found this superior to the more conventional L2 loss.

\subsection{Recourse Generation}

To generate recourse, we perform guided diffusion by adding an additional guiding term $\nabla_{\mathbf{z}_{l+1, t}} \cL_\text{guide}(\mathbf{z}_{l+1, t}, y^*)$ at each denoising step. Here, $\nabla_{\mathbf{z}_{l+1, t}} \cL_\text{guide}(\mathbf{z}_{l+1, t}, y^*)$ is the gradient of the guiding loss with respect to $\mathbf{z}_{l+1, t}$, and $y^*$ is the desired decision label. In this work, we define the guiding loss as follows: $\smash{\cL_\text{guide}(\bz, y^*) = \text{CE}\left(f(\bz), y^*\right)}$, where $\text{CE}$ denotes cross-entropy. 

\section{Datasets}
\label{sec:datasets}
We evaluate the effectiveness of our approach on 3 datasets: LoanSynth dataset, GradSynth dataset and AdultIncome++. Here, LoanSynth and GradSynth are synthetic datasets while AdultIncome++ is a semi-realistic dataset.

\subsection{LoanSynth Dataset}

\begin{table*}[t]
    \centering
    \setlength{\tabcolsep}{3.6pt}
    \small
    \caption{\textbf{LoanSynth Dataset Example.} We provide a sample trajectory of a user in our proposed synthetic LoanSynth dataset. Each row corresponds to a year in the user's life. Each column corresponds to an attribute of the user state. The last column corresponds to the decision-maker's label about whether this user would be approved for a loan or not.}
    \label{tab:loansynth_sample}
    \begin{tabular}{c|cccccccccc|c}
        \toprule
        $t$ & \textbf{Age} & \textcolor{teal}{\textbf{Edu}} & \textcolor{teal}{\textbf{Occupation}} & \textbf{Cash} & \textbf{Bank Bal} & \textcolor{teal}{\textbf{Life}} & \textbf{Cred} & \textcolor{teal}{\textbf{Wt}} & \textbf{Race} & \textbf{Sex} & \textbf{Label} \\
        \midrule
        0 & 18 & \textcolor{teal}{0} & \textcolor{teal}{\texttt{stud\_bachelor 1}} & -58K & -57K & \textcolor{teal}{L} & 300 & \textcolor{teal}{46.05} & C & M & Loan Denied \\
        1 & 19 & \textcolor{teal}{0} & \textcolor{teal}{\texttt{stud\_bachelor 2}} & -59K & -116K & \textcolor{teal}{L} & 300 & \textcolor{teal}{44.12} & C & M & Loan Denied \\
        2 & 20 & \textcolor{teal}{0} & \textcolor{teal}{\texttt{stud\_bachelor 3}} & -57K & -174K & \textcolor{teal}{L} & 300 & \textcolor{teal}{43.57} & C & M & Loan Denied \\
        3 & 21 & \textcolor{teal}{1} & \textcolor{teal}{\texttt{engineer 0}} & 37K & -136K & \textcolor{teal}{T} & 300 & \textcolor{teal}{41.68} & C & M & Loan Denied \\
        4 & 22 & \textcolor{teal}{1} & \textcolor{teal}{\texttt{engineer 1}} & 38K & -97K & \textcolor{teal}{T} & 300 & \textcolor{teal}{40.52} & C & M & Loan Denied \\
        5 & 23 & \textcolor{teal}{1} & \textcolor{teal}{\texttt{engineer 2}} & 42K & -55K & \textcolor{teal}{T} & 300 & \textcolor{teal}{42.31} & C & M & Loan Denied \\
        \bottomrule
    \end{tabular}
\end{table*}

In LoanSynth dataset, a user trajectory is a table with columns such as age, education level, occupation, cashflow, bank balance, lifestyle, credit score, cars owned, weight, race, gender, and a label column with respect to a decision-maker about whether the loan was approved for this user or not.  Each row corresponds to a time step or a year in the life of this user. Therefore, as we go down from the top row to the bottom row of the table, we can see the evolution of the state-of-the-user over time. We provide 30 rows or 30 years' worth of trajectory per user. 

To generate the data, we define an initial distribution of the user state and causal dynamics for the evolution of the user state. For more details about these definitions, see Appendix. For an example of this dataset, see Table \ref{tab:loansynth_sample}.

\subsection{GradSynth: A Synthetic Dataset for Graduate School Applications}

In GradSynth dataset, a user trajectory is a sequence of rows where the fields describe a student's state as they apply for college admissions: 1) the tier of their previous college, 2) GPA, 3) GRE, 4) work experience, 5) number of recommendation letters, 6) quality of their SOP, and 7) gender and a label column that specifies whether the user or the student was admitted to a graduate school or not. As we go down the rows of the table, the state of the user evolves. We provide 5 rows of trajectory per user. For a sample of this dataset, see Table \ref{tab:gradsynth_sample}.

\begin{table*}[t]
\small
\setlength{\tabcolsep}{4pt}
\caption{\textbf{GradSynth Dataset Example.} We provide a sample trajectory (with $t$ as the time index) of a user in the GradSynth dataset. Each column corresponds to an attribute of the user state. The \textcolor{teal}{teal-colored} columns are those that can be directly intervened upon by the user. The last column corresponds to the label about whether this user would be admitted to college or not.}
\centering
\begin{tabular}{@{}c|cccccccc|c@{}}
\toprule
$t$ & \textbf{Attempt} & \textbf{College} & \textbf{GPA} & \textbf{GRE Score} & \textcolor{teal}{\textbf{\texttt{\#}Exp}} & \textcolor{teal}{\textbf{\texttt{\#}LORs}} & \textcolor{teal}{\textbf{SOP}} & \textbf{Gender} & \textbf{Label} \\
\midrule
0 & $1^\text{st}$ & Tier-3 & 3.0 & Good & \textcolor{teal}{2} & \textcolor{teal}{0}         & \textcolor{teal}{Good}      & Female & Not Admitted \\
1 & $2^\text{nd}$ & Tier-3 & 3.0 & Good & \textcolor{teal}{3} & \textcolor{teal}{1}         & \textcolor{teal}{Poor}      & Female & Not Admitted \\
2 & $3^\text{rd}$ & Tier-3 & 3.0 & Good & \textcolor{teal}{3} & \textcolor{teal}{1}         & \textcolor{teal}{Good}      & Female & Not Admitted \\
3 & $4^\text{th}$ & Tier-3 & 3.0 & Good & \textcolor{teal}{3} & \textcolor{teal}{1}         & \textcolor{teal}{Good}      & Female & Not Admitted \\
4 & $5^\text{th}$ & Tier-3 & 3.0 & Good & \textcolor{teal}{3} & \textcolor{teal}{1}         & \textcolor{teal}{Excellent} & Female & Admitted  \\
\bottomrule
\end{tabular}
\label{tab:gradsynth_sample}
\end{table*}

\begin{table*}[t]
\centering
\small
\setlength{\tabcolsep}{3.4pt}
\caption{\textbf{AdultIncome++ Dataset Example.} We provide a sample trajectory (with $t$ as the time index) of a user in the AdultIncome++ dataset. Each column corresponds to an attribute of the user state. The \textcolor{teal}{teal-colored} columns are those that can be directly intervened upon by the user. The last column corresponds to the label about whether this user has an income greater than \$50K or not.}
\begin{tabular}{c|ccccccccc|c}
\toprule
$t$ & \textbf{Age}  & \textcolor{teal}{\textbf{Education}}       & \textcolor{teal}{\textbf{Marital Stat.}}     & \textcolor{teal}{\textbf{Occupation}}         & \textcolor{teal}{\textbf{Relationship}}       & \textbf{Race} & \textbf{Sex} & \textbf{Hrs/Wk} & \textbf{Nat} & \textbf{Label} \\
\midrule
0 & 24     & \textcolor{teal}{Some-college} & \textcolor{teal}{Never-married} & \textcolor{teal}{Tech-support}    & \textcolor{teal}{Not-in-family} & W & F & 37 & US & 0 \\
1 & 25         & \textcolor{teal}{Some-college} & \textcolor{teal}{Never-married} & \textcolor{teal}{Exec-manager} & \textcolor{teal}{Not-in-family} & W & F & 37 & US & 0 \\
2 & 26     & \textcolor{teal}{Some-college} & \textcolor{teal}{Never-married} & \textcolor{teal}{Exec-manager} & \textcolor{teal}{Not-in-family} & W & F & 37 & US & 0 \\
3 & 27        & \textcolor{teal}{Assoc-voc}    & \textcolor{teal}{Never-married} & \textcolor{teal}{Exec-manager} & \textcolor{teal}{Not-in-family} & W & F & 37 & US & 1 \\
\bottomrule
\end{tabular}
\label{tab:adultincomesample}
\end{table*}

\subsection{Adult Income ++ Benchmark}

In Adult Income++ dataset, a user trajectory is a table with the following columns about an adult: 1) age, 2) workclass, 3) fnlwgt, 4) education, 5) marital-status, 6) occupation, 7) relationship, 8) race, 8) sex, 10) capital-gain, 11) capital-loss, 12) hours-per-week, 13)  native-country, and a label column that specifies if the income is above $>50K$ or $=<50K$. As we go down the rows, state of the user evolves. We provide 4 rows of trajectory per user. For an example, see Table \ref{tab:adultincomesample}.  

\section{Related Work}

\textbf{Action Recourse.}
Several works have pursued the problem of action recourse \citep{ustun2019actionable, Karimi2020AlgorithmicRU, Karimi2020AlgorithmicRF, dhurandhar2018explanations, Rasouli2021CARECA, Verma2020CounterfactualEA, Jiang2024RobustCE, tominaga2024reassessing, detoni2024personalized, yetukuri2023towards, esfahani2024preference, koh2025understanding, wang2023gamcoach, kancheti2025hare}. However, to our knowledge, none of the existing methods acquire a notion of cost associated to an action in a data-driven manner. Although there are several methods about counterfactual explanations \citep{wachter2017counterfactual, mothilal2020explaining, Gomez2020ViCEVC, face, madaan2024navigating, Hashemi2020PermuteAttackCE}. Although some have incentives to maintain proximity or diversity \citep{Cui2015OptimalAE, karimi2020model, ustun2019actionable, Rawal2020BeyondIR, Kanamori2020DACEDC}, none of them incorporate an explicit notion of cost to the user. Another line of works tackles the issue of robustness of the generated recourse or explanations \citep{Hamman2023RobustCE, Maragno2023FindingRO, Guyomard2023GeneratingRC, Jiang2023ProvablyRA, Leofante2023PromotingCR, Kayastha2024LearningAugmentedRA}, including dealing with corrupted data \citep{Kanamori2023AlgorithmicRW}. However, this is orthogonal to our focus. Future works may bring together our approach of data-driven recourse and the advances in robustness of action recourse.

\textbf{Diffusion in Tabular Data.}
There has been recent literature leveraging diffusion models for performing certain tasks in tabular domain. TabDDPM \citep{kotelnikov2023tabddpm} has been introduced as a diffusion framework for mixed-type tables, outperforming GAN/VAE baselines.   TabDiff \citep{shi2024tabdiff} proposed a Transformer-parameterized joint diffusion with feature-wise learnable noise schedules and classifier-free guidance for conditional imputation.   TimeAutoDiff \citep{suh2024timeautodiff} has been developed combining a VAE with latent diffusion for one-shot, non-autoregressive synthesis of multivariate time-series tables. \citep{wen2024diffimpute} presented DiffPuter, which integrates EM with diffusion for iterative missing-value imputation, TabADM \citep{zamberg2023tabadm} for unsupervised anomaly detection via diffusion-based density modeling underscoring how diffusion-LM-inspired architectures enable high-fidelity synthesis, imputation, and analytic tasks across diverse tabular domains.

\begin{figure*}[t]
    \centering
    \includegraphics[width=\textwidth]{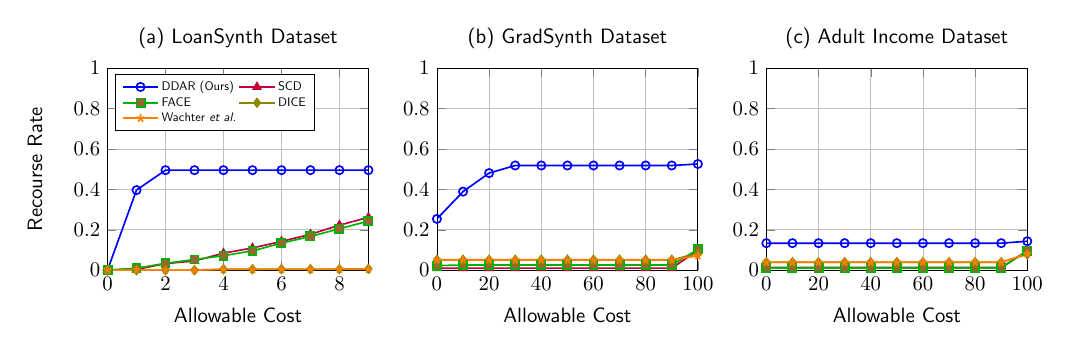}
    \caption{We plot the recourse rate as a function of maximum allowable cost for our method and compare it against the baselines on the datasets LoanSynth, GradSynth and AdultIncome++.}
    \label{fig:LS-ROC}
\end{figure*}

% RR-AUC/Cost and CNLL/UNLL placed side-by-side as a single wide float
\begin{table*}[t]
  \scriptsize
  \centering
  \sisetup{detect-weight=true,detect-family=true,group-digits=false,table-format=3.3}
  \begin{minipage}[t]{0.48\textwidth}
    \centering
    \caption{RR-AUC ($\uparrow$) and Cost ($\downarrow$) across the 3 datasets. Our model outperforms all baselines.}
    \label{tab:rrauc-cost}
    \begin{tabular}{lcccccc}
      \toprule
      {\textbf{Method}}
        & \multicolumn{2}{c}{\textbf{LoanSynth}}
        & \multicolumn{2}{c}{\textbf{GradSynth}}
        & \multicolumn{2}{c}{\textbf{AdultIncome++}} \\
      \cmidrule(lr){2-3}\cmidrule(lr){4-5}\cmidrule(lr){6-7}
        & {RR-AUC} & {Cost} & {RR-AUC} & {Cost} & {RR-AUC} & {Cost} \\
      \midrule
      Wachter et\! al. & 0.020 & 76.54 & 5.090 & 87.23 & 4.080 & 51.99 \\
      DiCE             & 0.026 & 79.68 & 5.207 & 88.38 & 4.097 & 49.99 \\
      FACE             & 0.885 & 107.0 & 2.750 & 222.1 & 5.758 & 98.93 \\
      SCD              & 0.943 & 68.51 & 1.256 & 227.7 & 5.965 & 86.48 \\ \midrule
      \textbf{Ours}    & \bfseries 4.142 & \bfseries 1.25
                       & \bfseries 48.98 & \bfseries 10.84
                       & \bfseries 13.38 & \bfseries 3.170 \\
      \bottomrule
    \end{tabular}
  \end{minipage}
  \hfill
  \begin{minipage}[t]{0.48\textwidth}
    \centering
    \caption{CNLL ($\downarrow$) and UNLL ($\downarrow$) across the 3 datasets. Our model outperforms all baselines.}
    \label{tab:cnll-unll}
    \sisetup{detect-weight=true,detect-family=true,group-digits=false,table-format=3.2}
    \begin{tabular}{lcccccc}
      \toprule
      {\textbf{Method}}
        & \multicolumn{2}{c}{\textbf{LoanSynth}}
        & \multicolumn{2}{c}{\textbf{GradSynth}}
        & \multicolumn{2}{c}{\textbf{AdultIncome++}} \\
      \cmidrule(lr){2-3}\cmidrule(lr){4-5}\cmidrule(lr){6-7}
        & {CNLL} & {UNLL} & {CNLL} & {UNLL} & {CNLL} & {UNLL} \\
      \midrule
      Wachter et\! al. & 75.29 & 59.87 & 61.35 & 46.28 & 53.20 & 25.23 \\
      DiCE             & 71.45 & 54.52 & 64.27 & 49.82 & 53.98 & 25.00 \\
      FACE             & 72.40 & 26.10 & 41.51 & 77.39 & 63.83 & 120.48 \\
      SCD              & 90.79 & 26.54 & 38.83 & 78.79 & 82.07 & 120.37 \\ \midrule
      \textbf{Ours}    & \bfseries 29.26 & \bfseries 21.27
                       & \bfseries 16.42 & \bfseries 17.56
                       & \bfseries 40.17 & \bfseries 19.50 \\
      \bottomrule
    \end{tabular}
  \end{minipage}
\end{table*}

\begin{figure*}[t]
    \centering
    \includegraphics[width=\textwidth]{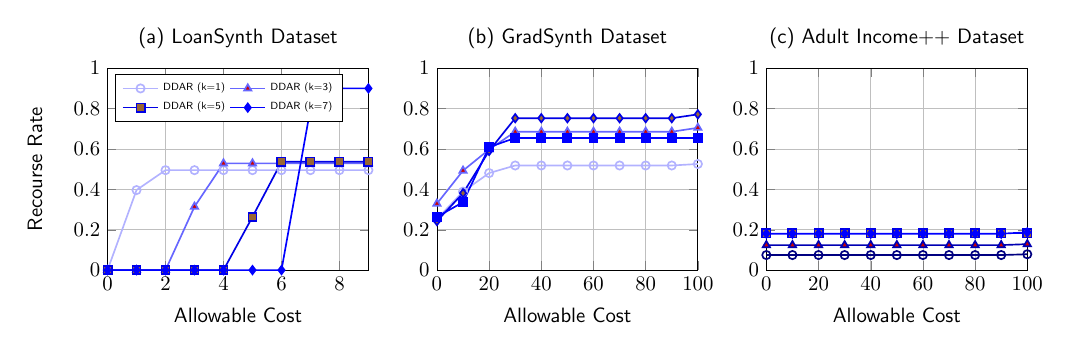}
    \caption{Recourse rate vs.\ steps $K$. \textit{Left:} LoanSynth, \textit{Centre:} GradSynth, \textit{Right:} AdultIncome++.}
    \label{fig:LS-ROC-kstep}
    \includegraphics[width=\textwidth]{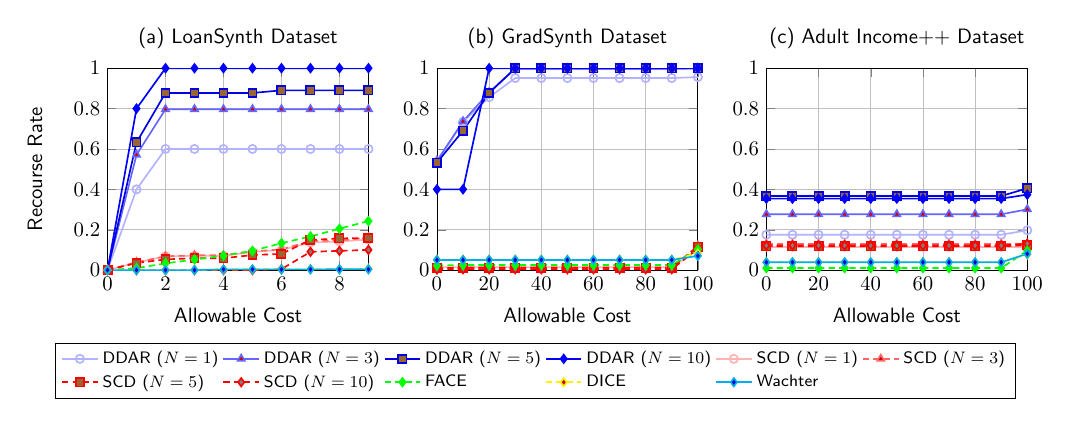}
    \caption{Recourse rate vs.\ number of trajectories $N$. \textit{Left:} LoanSynth, \textit{Centre:} GradSynth, \textit{Right:} AdultIncome++.}
    \label{fig:LS-ROC-nbranch}
\end{figure*}

\begin{table*}[t]
    \centering
    \small
    \setlength{\tabcolsep}{4pt}
    \caption{\textbf{Qualitative Comparison of Methods on LoanSynth Dataset.} We highlight all \cyan{changes} made by a method in cyan color. We also show \red{immutable} attributes which should not change like race and gender, and highlight the ones that were wrongly changed in red color.}
    \label{tab:qualitative_loansynth}
    \begin{tabular}{lccccccccccc}
        \toprule
        \textbf{Method} & \textbf{Age} & \textbf{Edu} & \textbf{Occ} & \textbf{Cash} & \textbf{Bank Bal} & \textbf{Spend} & \textbf{Debt} & \textbf{Cred} & \textbf{Race} & \textbf{Sex} & \textbf{Loan} \\
        \midrule
        {Input}  & 29 & 1 & ('sw\_dev', 8) & 30K & 83K & L & 629.36 & 75.06 & C & F & No \\ \noalign{\vskip 0.6ex} \cline{1-12} \noalign{\vskip 0.6ex}
        Ours   & \cyan{30} & 1 & \cyan{('sw\_dev', 9)} & \cyan{22K} & \cyan{127K} & L & \cyan{775.59} & \cyan{73.90} & C & F & Yes\\
        DICE    & 29 & \cyan{4} & \cyan{('scientist', 3)} & \cyan{30K} & \cyan{71K} & L & \cyan{627.35} & \cyan{74.83} & C & F & Yes \\
        SCD     & \cyan{36} & \cyan{2} & \cyan{('sw\_dev', 0)} & \cyan{3K} & \cyan{325K} & L & \cyan{849.99} & \cyan{79.58} & \red{H} & F & Yes \\
        FACE    & \cyan{42} & 1 & \cyan{('sw\_dev', 20)} & \cyan{23K} & \cyan{602K} & \cyan{T} & \cyan{849.99} & \cyan{90.88} & \red{H} & \red{M} & Yes \\
        Wachter & 29 & \cyan{4} & \cyan{('scientist', 3)} & \cyan{145K} & \cyan{71K} & L & \cyan{627.35} & \cyan{74.83} & C & F & Yes \\
        \bottomrule
    \end{tabular}
\end{table*}

\section{Experiments}

\subsection{Setup}
\subsubsection{Datasets}
We experiment on three datasets introduced in Section \ref{sec:datasets}: (1) LoanSynth, and (2) GradSynth and (3) Adult Income ++. In \textit{LoanSynth}, the decision model is tasked with predicting a loan decision (\textit{approve} or \textit{reject}) for a structured user state containing fields like \textit{cash flow}, or \textit{education} including immutable ones such as \textit{gender} and \textit{race}.  In \textit{GradSynth}, the decision model is tasked with predicting the admission decision (\textit{admit} or \textit{reject}). 
The \textit{Adult-Income++} is a synthetic extension of the original Adult-Income dataset \citep{frank2010uci} where we introduce certain dynamics on the user's attributes to evolve the user state over time. The decision maker is tasked with predicting whether the user's income is greater than \$50K or not.

\subsubsection{Metrics}
    \textbf{1. Recourse Rate.} We compute the fraction of all test instances for which the model could produce a recourse with the desired label under the given decision model. However, this alone would mean that a poor model can produce a recourse that can have an arbitrarily high cost. Therefore, we propose to report recourse rate that can achieve the desired label under a certain allowable cost.
    We plot this metric with respect to various values of the maximum allowable cost. The higher the recourse rate the better. Since the above is dependent on the maximum allowable cost, therefore, to obtain a single measure of comparison, we also report the area-under-curve for the recourse rate plot (or RR-AUC for short).\\
  \textbf{2. Cost.} We compute the average cost per generated recourse. The average is computed on all test instances. The cost is computed using a cost function defined by a human expert for each dataset. \\
    \textbf{3. Conditional Negative Log-Likelihood (CNLL).} As argued earlier, the likely next user state should also correspond to low-cost trajectories. Therefore, another metric of interest the log-likelihood of the model-generated recourses given the previous user state. To compute this, we utilize a conditional density model that is used to evaluate all the models' generated recourses given the previous user state.\\
   \textbf{4. Unconditional Negative Log-Likelihood (UNLL).} Another metric of interest is whether the generated recourse would make sense under any circumstance or not. Thus marginalizing over the conditioning in the conditional log-likelihood, we also report the unconditional log-likelihood. To compute this, we utilize a density model over the user states to evaluate the generated recourse from all models. 

\subsubsection{Baselines}
We compare DDAR with five baselines for generating recourses from structured datasets. We describe each of these baselines below:
\begin{enumerate}
    \item \textbf{Wachter et al. \citep{wachter2017counterfactual}} is an important baseline that tries to achieve the desired label while minimally perturbing the current user state.
    We implement this baseline by first encoding the user state and then applying gradient descent to search a recourse that produces the desired decision with smallest possible distance to the current user state. 
    \item \textbf{DiCE. \citep{mothilal2020explaining}} is another key baseline that, like Wachter et al., uses optimization to find a new user state which achieves the desired label with minimal perturbation and, importantly, with high diversity among various generated recourses. We implement this baseline by first encoding the row in a similar way as done in Wachter \citep{wachter2017counterfactual}. Next, gradient descent is applied to make find a set of recourses that achieve the desired decision, are in close proximity to the input sample, and are diverse enough. 
    \item \textbf{FACE. \citep{face}} This is a graph-based baseline for action recourse. It treats the given user state as a node in a graph and the edges capture the proximity between the nodes. Then, given an input, the recourse is generated by finding a low-distance path that crosses the decision boundary. This is the only baseline, to our knowledge, that tackles the question of feasibility of the trajectory, as it prevents the trajectory from passing through sparse regions. 
    \item \textbf{SCD \citep{madaan2024navigating}} is a counterfactual explainer that accounts for the underlying data distribution via a neural density modeling framework such as a VAE or a diffusion model but lacks a user trajectory model that can provide low-cost recourse recommendations.
\end{enumerate}

\subsection{Results}
\subsubsection{Can DDAR effectively provide low-cost recourse?}

\textbf{DDAR Provides High Recourse Rate.} To answer this, we look at the Recourse Rate plots in Figure \ref{fig:LS-ROC} and the values of RR-AUC reported in Table \ref{tab:rrauc-cost}. We can see that even with allowable cost of 2.0 on LoanSynth, our model has a recourse rate of 0.495 which is remarkably higher than all other baselines which remain close to 0. Similarly, in GradSynth, with maximum allowed cost of 20., other models are totally failing with recourse rate of almost 0.0 while our model achieves a much higher score of 0.518. This fact is also captured in our significantly higher RR-AUC score where we achieve 4.142 on LoanSynth while other baselines achieve a score less than 1.0. Similarly, on GradSynth, we achieve the RR-AUC of 48.98 while others have a score smaller than 6.0. We see a similar trend in Adult Income ++ dataset where our model achieves significantly higher RR-AUC of 4.142 while maintaining the lowest cost of 1.25 while other baselines achieve a RR-AUC score of less than 1.0.

\textbf{DDAR Provides Low Cost Recourse.} We also report the average cost associated to the generated recourses. We can see in all three datasets LoanSynth, GradSynth and Adult Income ++ that our model achieves a remarkably low cost while other models are significantly worse. This can be attributed to the fact that all the baselines (including ours) were given no explicit knowledge of the cost function and the models had to infer this from data. Other models, lacking a mechanism to infer this, are totally failing in this unsupervised regime, while ours does not.

\textbf{DDAR Provides Feasible Recourse.} There are two measures of feasibility. (1) The first measure is the conditional feasibility which measures how feasible is the recommended recourse given the state the user is currently in. We measure this via the conditional negative log-likelihood (CNLL) score and we can see that our recourses have a much lower CNLL of 29.26 on LoanSynth, 16.42 on GradSynth and 40.17 on Adult Income ++, while next best baseline achieves a much worse score of 71.45 on LoanSynth, 38.83 on GradSynth and 53.20 on Adult Income ++. (2) The second measure of feasibility is the unconditional feasibility which measures whether the recourse is coherent regardless of the current user state. In this measure also, our model achieves a score better than the other methods.
\textbf{Qualitative Analysis of the Generated Recourse.} We show the qualitative results for the LoanSynth dataset in Table \ref{tab:qualitative_loansynth}. Note that there are certain fields which should be immutable such as race or sex and changing them should be associated with a high cost. However, many baselines erroneously change race or sex in order to achieve the desired label. On the other hand, our model, through exposure to the historical data is able to take cognizance of this fact and produces recourse that do not change such immutable fields. Similarly, qualitative results for GradSynth and Adult Income ++ are presented in Appendix. 

\subsubsection{Benefits of Multi-Step Recourse}
A unique benefit of our method is that we can iteratively run the recourse generation process several times. Therefore, if no low-cost recourse is possible within a single time-step (e.g., a single year of the user's life), there may be recourse possible if more time is given to the user (e.g., a time horizon of multiple years of the user's life) to achieve their goal.
We report the recourse rates of $K$-step recourse with $K=1, 3, 5$ and $7$ in Figure \ref{fig:LS-ROC-kstep}. We see that as the $K$ is increased, we can achieve even higher rates of recourse. In LoanSynth dataset, we are able to achieve 0.9 recourse rate at $k=7$ at allowable cost of 8 while in GradSynth dataset we are able to reach to 0.80 recourse rate at allowable cost of 100 and in Adult Income we are able to achieve a recourse rate of ~0.24 at an allowable cost of 0.  Note that even with $K=1$, our model was already surpassing the other baselines. Qualitative $K$-step samples are shown in Table~\ref{tab:ddar_sample_K} in the Appendix.

\subsubsection{Best-of-$N$ Recourse.}
A benefit of our method is that we can sample multiple recourse trajectories and then choose the best one. We evaluate the effect of using various values of $N$ in Figure \ref{fig:LS-ROC-nbranch}. We note that across all datasets, a larger $N$ e.g., 5 or 10, leads to a boost in our recourse rate for any maximum allowable cost. In comparison, other baselines are either deterministic, thus incapable of deriving benefits of sampling, or the boost they show with greater $N$ is not as large as that of our model. For instance, in LoanSynth dataset, our model is able to achieve 1.0 recourse rate at an allowable cost of 2.0 while the next best baselines is at 0.12. Similar trend follows for GradSynth and Adult Income ++ datasets. 

\section{Conclusion}
In this paper, we introduced the first data-driven framework called Data-Driven Action Recourse (DDAR) for generating recourse without providing any costly common-sense knowledge from humans. We contribute three datasets in this work: LoanSynth, GradSynth and Adult Income++ for generating data-driven recourse. Through experiments, we show that DDAR generates recourses that can achieve the desired decision while also maintaining low cost compared to the baselines. An interesting avenue of future work is to have user-defined constraints that can be incorporated into the data-driven paradigm.

% References
% No \bibliographystyle here: aaai2027.sty already sets it when natbib is loaded.
\bibliography{ijcai26}

% -----------------------------------------------------------------------
% Technical Appendix
% -----------------------------------------------------------------------
\appendix
\section{Additional Details of the Method}

\begin{table*}[t]
    \centering
    \small
    \caption{\textbf{Qualitative Analysis of DDAR with $K$-step recourse on LoanSynth dataset.} We provide input-output pairs from our system DDAR. Each row corresponds to a user's data at a point in time. Each column corresponds to an attribute of the user's state. $K=10$.}
    \label{tab:ddar_sample_K}
    \begin{tabular}{ccccccccccc}
        \toprule
        \textbf{Age} & \textbf{Edu} & \textbf{Occ} & \textbf{Cash} & \textbf{Bank Bal} & \textbf{Life} & \textbf{Cred} & \textbf{Wt} & \textbf{Race} & \textbf{Sex} & \textbf{Label} \\
        \midrule
        28 & 1 & \texttt{soft\_dev 1} & 28K & 60K & T & 470.80 & 58.12 & B & F & No \\
        38 & 2 & \texttt{manager 0} & 24K & 353K & M & 849.99 & 61.55 & B & F & Yes \\
        38 & 1 & \texttt{soft\_dev 11} & 60K & 550K & T & 849.99 & 54.90 & B & F & Yes \\
        \midrule
        24 & 1 & \texttt{stud\_mast 1} & -58K & -266K & L & 300.00 & 82.90 & C & F & No \\
        34 & 3 & \texttt{artist 4} & 15K & -350K & M & 300.04 & 82.43 & C & F & Yes \\
        34 & 2 & \texttt{engineer 9} & 60K & 99K & M & 616.17 & 87.18 & C & F & Yes \\
        \bottomrule
    \end{tabular}
\end{table*}

\subsection{Denoiser Model and Training Hyperparameters}
In Table \ref{tab:hyperparams_denoiser}, we report the various hyperparameters that we used for training the diffusion model.
\begin{table}[!ht]
\centering
\begin{tabular}{@{}ll@{}}
\toprule
\textbf{Hyperparameter}               & \textbf{Value}       \\ 
\midrule
Learning Rate                         & 0.0001               \\
Learning Rate Warmup Steps            & 30000                \\
Learning Rate Decay Half-Life         & 250000               \\
Gradient Clipping                     & 0.05                 \\
Diffusion Steps                       & 2000                 \\
Batch Size                            & 120                  \\
\midrule
Dropout       & 0.1                  \\
Hidden Activation                          & GELU                 \\
Hidden Size                          & 768                  \\
Intermediate Size                   & 3072                 \\
Number of Attention Heads                & 12                   \\
Number of Layers                  & 12                   \\
Vocabulary Size                          & 30522                \\
\bottomrule
\end{tabular}
\caption{Hyperparameters of the Diffusion Model.}
\label{tab:hyperparams_denoiser}
\end{table}

\textbf{Schedule for $\beta$.} We use the cosine schedule for the denoising diffusion used in the existing works \citep{nichol2021improved}.

\subsection{Sinusoidal Time Embeddings}
The denoiser $g_\ta(\bz_{l+1,t}, \bz_{l}, t)$ in our diffusion model requires $t$ as an input. Since $t$ is a scalar, we embed it as a vector. For this, we adopt the sinusoidal embedding approach of previous works. Formally, we embed $t$ as $e(t)$ where:
\begin{align*}
    e(t) = [&\text{cos}(\omega_0 t), \ldots, \text{cos}(\omega_{H-1}t), \\
           &\text{sin}(\omega_0 t), \ldots, \text{sin}(\omega_{H-1} t)]
\end{align*}
Here, $H=D/2$ with $D$ as the dimension of the embedding; and $\omega_k=\text{exp}(-\log(\text{max\_period})k/H)$ with $\text{max\_period}=10000$.

\section{Cost Functions}
In this section, we describe the pseudo-code used for computing costs in our two benchmarks for evaluation purposes. For these, see Algorithms \ref{algo:cost_loansynth} and \ref{algo:cost_gradsynth}. These cost functions are human-defined. Note that these human-defined cost functions are not provided to any of our evaluated models---all of which were trained in an unsupervised manner on historical user log data. 

\begin{algorithm}
\caption{Algorithm to Compute Cost for the LoanSynth Dataset}
\label{algo:cost_loansynth}
\begin{algorithmic}[1]
\Function{LoanSynthCost}{\texttt{state\_cur}, \texttt{state\_next}}
    \State \texttt{cost} $\gets 0$
    \State
    \State \text{\# The age can never decrease.}
    \If{\texttt{state\_cur[`age']} $\ge$ \texttt{state\_next[`age']}}
        \State \texttt{cost} $\gets$ \texttt{cost} + 100.0
    \EndIf
    \State
    \State \text{\# The more the user has to wait, the more the cost.}
    \State \texttt{cost}  $\gets$ \texttt{cost}  + |\texttt{state\_cur[`age']} - \texttt{state\_next[`age']}|
    \State
    \State \text{\# Lifestyle change has a cost.}
    \If{\texttt{state\_cur[`lifestyle']} $\neq$ \texttt{state\_next[`lifestyle']}}
        \State \texttt{cost}  $\gets$ \texttt{cost}  + 1.0
    \EndIf
    \State   
    \State \text{\# Education-level cannot go down with time.}
    \If{\texttt{state\_cur[`lifestyle']} $>$ \texttt{state\_next[`lifestyle']}}
        \State \texttt{cost} $\gets$ \texttt{cost} + 100.0
    \EndIf
    \State
    \State \Return \texttt{cost}
\EndFunction
\end{algorithmic}
\end{algorithm}

\begin{algorithm}
\caption{Algorithm to Compute Cost for the GradSynth Dataset}
\label{algo:cost_gradsynth}
\begin{algorithmic}[1]
\Function{CustomCost}{\texttt{state\_cur}, \texttt{state\_next}}
    \State \texttt{cost} $\gets 0$
    \State
    \State \text{\# University Tier should not change}
    \If{\texttt{state\_cur[`tier']} $\neq$ \texttt{state\_next[`tier']}}
        \State \texttt{cost} $\gets$ \texttt{cost} + 100.0
    \EndIf

    \State
    \State \text{\# GPA should not change}
    \If{\texttt{state\_cur[`gpa']} $\neq$ \texttt{state\_next[`gpa']}}
        \State \texttt{cost} $\gets$ \texttt{cost} + 100.0
    \EndIf

    \State
    \State \text{\# Recommendations should increase}
    \If{\texttt{float}(\texttt{state\_cur[`recommendation']}) > \texttt{float}(\texttt{state\_next[`recommendation']})}
        \State \texttt{cost} $\gets$ \texttt{cost} + 100.0
    \EndIf

    \State
    \State \text{\# Experience Letters cannot decrease}
    \If{\texttt{float}(\texttt{state\_cur[`experience']}) > \texttt{float}(\texttt{state\_next[`experience]})}
        \State \texttt{cost} $\gets$ \texttt{cost} + 100.0
    \EndIf

    \State
    \State \text{\# Gender should not change}
    \If{\texttt{state\_cur[`gender']} $\neq$ \texttt{state\_next[`gender']}}
        \State \texttt{cost} $\gets$ \texttt{cost} + 100.0
    \EndIf

    \State
    \State \text{\# GRE increase comes with a cost}
    \State \texttt{cost} $\gets$ \texttt{cost} + $10.0 \times \texttt{relu}(\texttt{state\_next[`gre']} - \texttt{state\_cur[`gre']})$

    \State
    \State \text{\# SOP quality increase comes with a cost}
    \State \texttt{cost} $\gets$ \texttt{cost} + $10.0 \times \texttt{relu}(\texttt{state\_next[`sop']} - \texttt{state\_cur[`sop']})$

    \State
    \State \Return \texttt{cost}
\EndFunction
\end{algorithmic}
\end{algorithm}

\section{Benchmark Datasets}
\subsection{LoanSynth Generation}
We show the data generation process for LoanSynth dataset in Algorithm \ref{algo:data_loansynth}. 

\begin{algorithm*}
\caption{Synthetic Loan Approval Data Generation}
\label{algo:data_loansynth}
\begin{algorithmic}[1]
\State \textbf{Input:} $N$ (number of individuals), $T$ (simulation duration in years)
\State \textbf{Output:} Synthetic dataset of individuals over time

\For{$i = 1$ to $N$}
    \State \textbf{Initialize state:}
    \State $age \gets \text{initial\_age()}$, \quad $education \gets \text{initial\_education()}$
    \State $occupation \gets \text{initial\_occupation()}$, \quad $cashflow \gets 0$
    \State $bank\_balance \gets \text{initial\_bank\_balance()}$, \quad $lifestyle \gets \text{initialize\_lifestyle()}$
    \State $credit\_score \gets \text{initial\_credit\_score()}$, \quad $weight \gets \text{initialize\_weight()}$
    \State $race \gets \text{initialize\_race()}$, \quad $gender \gets \text{initialize\_gender()}$, \quad $loan\_approved \gets 0$
    
    \State $S \gets []$ \Comment{Initialize storage for state history}

    \For{$t = 1$ to $T$}
        \State $age \gets \text{transition\_age}(age)$
        \State $(occupation, education) \gets \text{transition\_occupation\_education}(occupation, education)$
        \State $lifestyle \gets \text{transition\_lifestyle}(lifestyle)$
        \State $weight \gets weight + \text{random}(-2,2)$
        \State $cashflow \gets \text{transition\_cashflow}(occupation, lifestyle)$
        \State $bank\_balance \gets \text{transition\_bank\_balance}(bank\_balance, cashflow)$
        \State $credit\_score \gets \text{transition\_credit\_score}(credit\_score, cashflow, bank\_balance)$
        \State $loan\_approved \gets \text{loan\_approval}(credit\_score, cashflow, bank\_balance)$
        \State $S \gets S \cup \{(age, education, occupation, lifestyle, weight, bank\_balance, credit\_score, loan\_approved)\}$
    \EndFor

    \State Save $S$ as CSV: \texttt{synthetic\_loan\_approval/person\_data\_i.csv}
\EndFor

\end{algorithmic}
\end{algorithm*}

\subsection{GradeSynth Generation}
We show the data generation process for GradSynth dataset in Algorithm \ref{algo:data_gradsynth}.

\begin{algorithm*}
\caption{Graduate Admissions Data Generation}
\label{algo:data_gradsynth}
\begin{algorithmic}[1]
\State \textbf{Input:} $N$ (number of applicants), $T$ (simulation duration in years)
\State \textbf{Output:} Synthetic dataset of graduate admission decisions

\For{$i = 1$ to $N$}
    \State \textbf{Initialize state:}
    \State $college \gets \text{initial\_college()}$, \quad $gpa \gets \text{initial\_gpa()}$
    \State $gre \gets \text{initial\_gre()}$, \quad $(experience, recommendation) \gets \text{initial\_experience\_recommendation()}$
    \State $sop \gets \text{initial\_sop()}$, \quad $gender \gets \text{initial\_gender()}$
    \State $admission \gets \text{None}$

    \State $S \gets []$ \Comment{Initialize storage for state history}

    \For{$t = 1$ to $T$}
        \State $(experience, recommendation) \gets \text{transition\_experience\_recommendation}(experience, recommendation)$
        \State $college \gets \text{transition\_college}(college)$
        \State $gpa \gets \text{transition\_gpa}(gpa)$
        \State $gre \gets \text{transition\_gre}(gre)$
        \State $sop \gets \text{transition\_sop}(sop)$
        \State $gender \gets \text{transition\_gender}(gender)$
        \State $admission \gets \text{compute\_admission\_decision}(college, gpa, gre, experience, recommendation, sop, gender)$
        \State $S \gets S \cup \{(college, gpa, gre, experience, recommendation, sop, gender, admission)\}$
    \EndFor

    \State Save $S$ as CSV: \texttt{synthetic\_grad\_admissions\_03/grad\_data\_i.csv}
\EndFor

\end{algorithmic}
\end{algorithm*}

\subsection{Adult Income ++ Generation}
We show the data generation process for Adult Income++ dataset in Algorithm \ref{algo:data_adult}.

\begin{algorithm}
\caption{Synthetic Adult Income Data Generation}
\label{algo:data_adult}
\begin{algorithmic}[1]
\Require Transition matrices $T_a$ for each $a\in\{\text{workclass},\dots,\text{sex}\}$  
\Require Legality rules $\mathcal{R}$; income model; original file \texttt{adult\_income.csv}  
\Ensure Generated sequences in \texttt{recourses/}
\State Load CSV, drop \texttt{education-num}, strip whitespace, remove rows with '?' in any categorical column
\State Create directory \texttt{recourses/} if needed
\ForAll{row in data}
  \State $state \gets \text{row[categorical]},\; age \gets \text{row.age},\; outputs \gets \{row\}$
  \For{$i=1\to4$}
    \State age$\gets$age+1
    \Repeat 
      \ForAll{$a$ in categorical} 
        \State $state[a]\sim T_a[state[a]]$
      \EndFor
    \Until{$state$ satisfies $\mathcal{R}$}
    \State $income\gets$ income\_model.predict(state)
    \State Append to outputs the record with updated $(state,age,income)$ and unchanged other fields
  \EndFor
  \State Write outputs to \texttt{recourses/sample\_\{row\_idx\}.csv}
\EndFor
\end{algorithmic}
\end{algorithm}

\section{Qualitative Samples}
In this section, we show additional qualitative samples generated by our framework DDAR and by baselines for LoanSynth in Table \ref{tab:qualloan} and for GradSynth in Table \ref{tab:qualgrad}.

\begin{table*}[h]
    \centering
    \small
    \renewcommand{\arraystretch}{1.2} % Adjust row spacing
    \setlength{\tabcolsep}{3pt}      % Tighten column spacing
    \caption{\textbf{Qualitative Comparison of Methods.} Changes in cyan, invalid immutable changes in red.}
    \label{tab:qualloan}
    \begin{tabular}{l r r l r r l r r c c c}
        \toprule
        Method & Age & Edu. & Occ. & Sav.\,(k) & Inc.\,(k) & Spd. & Debt\,(k) & CrSc. & Race & Sex & Ln. \\
        \midrule
        Input   & 29 & 1 & SD-8  & 30.4  & 83.3   & lav.  & 0.63 & 75.1 & C & F & No  \\
        DSCD    & 30 & 1 & SD-9  & 21.8  & 127.5  & lav.  & 0.78 & 73.9 & C & F & Yes \\
        DICE    & 29 & 4 & SC-3  & 30.3  & 71.2   & lav.  & 0.63 & 74.8 & C & F & Yes \\
        SCD     & 36 & 2 & SD-0  & 2.8   & 325.1  & lav.  & 0.85 & 79.6 & \red{H} & F & Yes \\
        FACE    & 42 & 1 & SD-20 & 22.8  & 602.1  & thr.  & 0.85 & 90.9 & \red{H} & \red{M} & Yes \\
        Wac.    & 29 & 4 & SC-3  & 145.2 & 71.2   & lav.  & 0.63 & 74.8 & C & F & Yes \\
        \midrule

        Input   & 33 & 0 & SB-3  & -29.3 & -88.3  & thr.  & 0.52 & 54.6 & B & M & No  \\
        DSCD    & 34 & 1 & SD-0  & 28.2  & -68.4  & thr.  & 0.85 & 56.8 & B & M & Yes \\
        DICE    & 33 & 0 & SB-3  & -29.0 & -96.9  & thr.  & 0.52 & 54.9 & B & M & Yes \\
        SCD     & 36 & 2 & SD-0  & 2.8   & 325.1  & lav.  & 0.85 & 79.6 & \red{H} & \red{F} & Yes \\
        FACE    & 38 & 1 & EN-1  & 53.2  & 428.9  & thr.  & 0.85 & 90.1 & \red{H} & M & Yes \\
        Wac.    & 33 & 0 & SB-3  & -29.0 & -96.9  & thr.  & 0.52 & 54.9 & B & M & Yes \\
        \midrule

        Input   & 29 & 1 & SD-2  & 33.7  & 93.4   & thr.  & 0.60 & 58.5 & B & F & No  \\
        DSCD    & 30 & 1 & SD-3  & 40.8  & 127.5  & thr.  & 0.78 & 56.8 & B & F & Yes \\
        DICE    & 29 & 4 & SC-3  & 34.5  & 99.4   & thr.  & 0.60 & 58.7 & B & F & Yes \\
        SCD     & 34 & 1 & SD-11 & 40.8  & 184.0  & mod.  & 0.85 & 75.8 & B & \red{M} & Yes \\
        FACE    & 38 & 1 & EN-1  & 53.2  & 428.9  & thr.  & 0.85 & 90.1 & \red{H} & \red{M} & Yes \\
        Wac.    & 29 & 4 & SC-3  & 145.2 & 99.4   & thr.  & 0.60 & 58.7 & B & F & Yes \\
        \midrule

        Input   & 32 & 1 & SM-1  & -39.4 & 191.7  & mod.  & 0.85 & 81.1 & A & F & No  \\
        DSCD    & 33 & 2 & EN-0  & 28.2  & 212.0  & mod.  & 0.85 & 82.4 & A & F & Yes \\
        DICE    & 32 & 1 & SC-3  & -39.5 & 184.0  & mod.  & 0.85 & 81.5 & A & F & Yes \\
        SCD     & 34 & 1 & SD-11 & 40.8  & 184.0  & mod.  & 0.85 & 75.8 & A & \red{M} & Yes \\
        FACE    & 41 & 1 & SD-2  & 61.5  & 528.1  & thr.  & 0.85 & 76.9 & \red{H} & \red{M} & Yes \\
        Wac.    & 32 & 1 & MG-16 & -39.5 & 184.0  & mod.  & 0.85 & 81.5 & A & F & Yes \\
        \midrule

        Input   & 35 & 2 & MG-4  & 32.1  & 95.8   & lav.  & 0.61 & 75.8 & H & M & No  \\
        DSCD    & 36 & 2 & MG-5  & 30.3  & 127.5  & lav.  & 0.78 & 74.8 & H & M & Yes \\
        DICE    & 35 & 4 & SC-3  & 145.2 & 99.4   & lav.  & 0.61 & 75.8 & H & M & Yes \\
        SCD     & 41 & 2 & SD-0  & 4.9   & 184.0  & lav.  & 0.85 & 63.4 & H & M & Yes \\
        FACE    & 41 & 2 & SD-2  & 61.5  & 528.1  & thr.  & 0.85 & 76.9 & H & M & Yes \\
        Wac.    & 35 & 4 & SP-1  & 145.2 & 99.4   & lav.  & 0.61 & 75.8 & H & M & Yes \\
        \bottomrule
    \end{tabular}
\end{table*}

\begin{table*}[h]
    \centering
    \small
    \renewcommand{\arraystretch}{1.6} % Adjust row spacing
    \setlength{\tabcolsep}{6pt} % Adjust column spacing
    \caption{Comparison of Different Methods on GradSynth Dataset. We show highlight all \cyan{changes} made my different algorithms in cyan color. We also show \red{immutable} attributes which should not change in a plausible recourse like gender, tier of institute, college gpa and highlight the ones that were changed in red color.}
    \label{tab:qualgrad}
    \begin{tabular}{lcccccccc}
        \toprule
        \textbf{Method} & \textbf{College} & \textbf{GPA} & \textbf{GRE} & \textbf{Exp.} & \textbf{Rec.} & \textbf{SOP} & \textbf{Gender} & \textbf{Admission} \\
        \midrule
        \textbf{Input}  & Tier-3 & 4.0 & Good    & 1.0 & 0.0 & Average   & Male   & False \\
        DSCD (ours)   & Tier-3 & 4.0 & Good    & 1.0 & 0.0 & \cyan{Excellent} & Male   & True \\
        SCD     & Tier-3 & 4.0 & \cyan{Poor}    & \cyan{8.0} & 0.0 & \cyan{Good}      & \red{Female} & True \\
        DICE    & \red{Tier-1} & 4.0 & \cyan{Excellent} & \cyan{8.0} & \cyan{6.0} & \cyan{Excellent} & Male   & True \\
        Wachter & \red{Tier-1} & 4.0 & \cyan{Poor}    & \cyan{8.0} & \cyan{6.0} & \cyan{Excellent} & Male   & True \\
        FACE    & Tier-3 & 4.0 & \cyan{Poor}    & \cyan{7.0} & \cyan{4.0} & Average   & Male   & True \\
        \midrule
        \textbf{Input}  & Tier-2 & 3.0 & Average & 0.0 & 0.0 & Excellent & Female & False \\
        DSCD (ours)   & Tier-2 & 3.0 & Average & \cyan{1.0} & 0.0 & Excellent & Female & True \\
        SCD     & Tier-3 & \cyan{4.0} & \cyan{Poor}    & \cyan{8.0} & \cyan{6.0} & \cyan{Average}   & Female & True \\
        DICE    & \red{Tier-1} & \cyan{4.0} & \cyan{Excellent} & \cyan{8.0} & \cyan{6.0} & Excellent & Female & True \\
        Wachter & \red{Tier-1} & \red{4.0} & \cyan{Poor}    & \cyan{8.0} & \cyan{6.0} & Excellent & Female & True \\
        FACE    & \red{Tier-3} & \red{4.0} & \cyan{Poor}    & \cyan{7.0} & \cyan{4.0} & \cyan{Average}   & Female & True \\
        \midrule
        \textbf{Input}  & Tier-3 & 4.0 & Average & 1.0 & 0.0 & Average   & Female & False \\
        DSCD (ours)   & Tier-3 & 4.0 & Average & \cyan{2.0} & 0.0 & \cyan{Good}      & Female & True \\
        SCD     & Tier-3 & 4.0 & \cyan{Poor}    & \cyan{8.0} & \cyan{6.0} & Average   & Female & True \\
        DICE    & \red{Tier-1} & 4.0 & \cyan{Excellent} & \cyan{8.0} & \cyan{6.0} & \cyan{Excellent} & Female & True \\
        Wachter & \red{Tier-1} & 4.0 & \cyan{Poor}    & \cyan{8.0} & \cyan{6.0} & \cyan{Excellent} & Female & True \\
        FACE    & Tier-3 & 4.0 & \cyan{Poor}    & \cyan{7.0} & \cyan{2.0} & Average   & \red{Male}   & True \\
        \midrule
        \textbf{Input}  & Tier-3 & 3.0 & Average & 0.0 & 0.0 & Good      & Female & False \\
        DSCD (ours)    & Tier-3 & 3.0 & Average & 0.0 & \cyan{6.0} & Good      & Female & True \\
        SCD     & Tier-3 & \red{4.0} & \cyan{Poor}    & \cyan{8.0} & \cyan{1.0} & \cyan{Average}   & Female & True \\
        DICE    & Tier-3 & \red{4.0} & \cyan{Poor}    & \cyan{8.0} & \cyan{6.0} & \cyan{Poor}      & Female & True \\
        Wachter & Tier-3 & \red{4.0} & \cyan{Poor}    & \cyan{8.0} & \cyan{6.0} & \cyan{Poor}      & Female & True \\
        FACE    & Tier-3 & \red{4.0} & \cyan{Poor}    & \cyan{7.0} & \cyan{4.0} & \cyan{Average}   & \red{Male}   & True \\
        \bottomrule
    \end{tabular}
\end{table*}

\begin{table*}[t]
    \centering
    \caption{\textbf{Qualitative Analysis of System DDAR.} We provide input-output pairs from our system DDAR. Each row corresponds to a user's data at a point in time. Each column corresponds to an attribute of the user's state.}
    \vspace{1em}
    \label{tab:ddar_sample}
    \begin{tabular}{ccccccccccc}
        \toprule
        \textbf{Age} & \textbf{Edu} & \textbf{Occ} & \textbf{Cash} & \textbf{Bank Bal} & \textbf{Life} & \textbf{Cred} & \textbf{Wt} & \textbf{Race} & \textbf{Sex} & \textbf{Loan?} \\
        \midrule
        29 & 1 & \texttt{soft\_dev 8} & 30,389 & 83,330.26 & lavish & 629.36 & 75.06 & C & F & No \\
        
        30 & 1 & \texttt{soft\_dev 9} & 21,816.25 & 127,471.26 & lavish & 775.59 & 73.90 & C & F & Yes \\
        \midrule
        33 & 0 & \texttt{stud\_bach 3} & -29,344 & -88,282.86 & thrifty & 523.62 & 54.60 & B & M & No \\
        
        34 & 1 & \texttt{soft\_dev 0} & 28,183.54 & -68,442.25 & thrifty & 849.99 & 56.80 & B & M & Yes \\
        \midrule
        29 & 1 & \texttt{soft\_dev 2} & 33,652 & 93,366.83 & thrifty & 597.82 & 58.49 & B & F & No \\
        
        30 & 1 & \texttt{soft\_dev 3} & 40,846.97 & 127,471.26 & thrifty & 775.59 & 56.80 & B & F & Yes \\
        \midrule
        32 & 1 & \texttt{stud\_mast 1} & -39,448 & 191,681.03 & moderate & 850.00 & 81.06 & A & F & No \\
        
        33 & 2 & \texttt{engineer 0} & 28,183.54 & 212,039.41 & moderate & 849.99 & 82.43 & A & F & Yes \\
        \midrule
        35 & 2 & \texttt{manager 4} & 32,057 & 95,828.00 & lavish & 607.82 & 75.79 & H & M & No \\
        
        36 & 2 & \texttt{manager 5} & 30,282.50 & 127,471.26 & lavish & 775.59 & 74.83 & H & M & Yes \\
        \bottomrule
    \end{tabular}
\end{table*}

\begin{table*}[ht!]
    \centering
    \small
    \setlength{\tabcolsep}{8pt} % Adjust column spacing
    \caption{\textbf{Qualitative Comparison of Methods on GradSynth Dataset.} We highlight all \cyan{changes} made by a method in cyan color. We also show \red{immutable} attributes which should not change in a plausible recourse like gender, tier of institute, college GPA and highlight the ones that were wrongly changed in red color. We see that other baselines often make such mistakes while ours does not.}
    \label{tab:qualitative_gradsynth}
    \vspace{1em}
    \begin{tabular}{lcccccccc}
        \toprule
        \textbf{Method} & \textbf{College} & \textbf{GPA} & \textbf{GRE} & \textbf{Exp.} & \textbf{Rec.} & \textbf{SOP} & \textbf{Gender} & \textbf{Admission} \\
        \midrule
        \textbf{Input}  & Tier-2 & 2.0 & Average & 2.0 & 0.0 & Good      & Female & False \\ \noalign{\vskip 0.6ex} \cline{1-9} \noalign{\vskip 0.6ex}
        Ours   & Tier-2 & 2.0 & Average & 2.0 & \cyan{6.0} & \cyan{Excellent} & Female & True \\
        SCD     & \red{Tier-3} & \cyan{4.0} & \cyan{Poor}    & \cyan{8.0} & \cyan{6.0} & Average   & Female & True \\
        DICE    & \red{Tier-1} & \cyan{4.0} & \cyan{Poor}    & \cyan{8.0} & \cyan{6.0} & \cyan{Excellent} & Female & True \\
        Wachter & \red{Tier-1} & \cyan{4.0} & \cyan{Poor}    & \cyan{8.0} & \cyan{6.0} & \cyan{Excellent} & Female & True \\
        FACE    & \red{Tier-3} & \cyan{4.0} & \cyan{Poor}    & \cyan{7.0} & \cyan{4.0} & Average   & Female & True \\
        \bottomrule
    \end{tabular}
\end{table*}

\end{document}